# A review of graph neural network applications in mechanics-related domains

Yingxue Zhao [1], Haoran Li [1], Haosu Zhou [1], Hamid Reza Attar [1], Tobias Pfaff [2], Nan Li [1*]

[1] Dyson School of Design Engineering, Imperial College London, London, UK
[2] Google DeepMind, London, UK
* Corresponding author. E-mail address: n.li09@imperial.ac.uk (N. Li)

**Abstract**

Mechanics-related problems often present unique challenges in achieving accurate geometric and physical representations, particularly for non-uniform structures. Graph neural networks (GNNs) have emerged as a promising tool to tackle these challenges by adeptly learning from graph data with irregular underlying structures. Consequently, recent years have witnessed a surge in complex mechanics-related applications inspired by the advancements of GNNs. Despite this process, there is a notable absence of a systematic review addressing the recent advancement of GNNs in solving mechanics-related problems. To bridge this gap, this review article aims to provide an in-depth overview of the GNN applications in mechanics-related domains while identifying key challenges and outlining potential future research directions. In this review article, we begin by introducing the fundamental algorithms of GNNs that are widely employed in mechanics-related applications. We provide a concise explanation of their underlying principles to establish a solid understanding that will serve as a basis for exploring the applications of GNNs in mechanics-related domains. The scope of this paper is intended to cover the categorisation of literature into solid mechanics, fluid mechanics, and interdisciplinary mechanics-related domains, providing a comprehensive summary of graph representation methodologies, GNN architectures, and further discussions in their respective subdomains. Additionally, open data and source codes relevant to these applications are summarised for the convenience of future researchers. This article promotes an interdisciplinary integration of GNNs and mechanics and provides a guide for researchers interested in applying GNNs to solve complex mechanics-related problems.

# 1. Introduction

## 1.1 Preamble to machine learning approaches in mechanics-related domains

Mechanics, as a fundamental discipline in the physical sciences, plays a pivotal role in various applications, for example, solid mechanics, fluid mechanics, and interdisciplinary mechanics-related domains. Solid mechanics delves into the motion and deformation of solid materials under external forces (Wang and Qin 2020). Fluid mechanics investigates the behaviours of fluids in stationary or moving states. Interdisciplinary mechanics-related applications tackle complex or multi-disciplinary systems drawing upon mechanics principles alongside other domains such as earth science, biology, or thermodynamics.

The mechanics-related domains often involve complex mathematical descriptions due to nonlinear, multiscale and multiphysics coupling phenomena (Groen et al. 2012; Lebon and Ramière 2023). Consequently, deriving analytical solutions for these mathematical models that apply to real-world scenarios becomes challenging. Conventionally, experiments are often conducted to provide real-world observation of physical behaviours, offering insights into new phenomena, and devise practical solutions to complex problems. However, experimental approaches are often resource-expensive and time-consuming. To mitigate this, numerical simulations provide a cost-effective means to address complex challenges, enabling iterative assessments without the need for costly experimental prototyping trials and offering the potential for parallel execution (Behbahani et al. 2009; Pietrzyk 2000; Pinto et al. 2017; Sharma and Kalamkar 2016). While numerical simulations have long played a pivotal role in offering robust and validated methods to solve mechanics-related tasks, they come with their own set of challenges, such as high computational costs and numerical expertise requirements.

Machine learning (ML) has emerged as a powerful complementary approach to experiments and simulations, significantly improving the efficiency of handling mechanics-related tasks. These ML models provide valuable insights, aiding in navigating complex parameter spaces inherent in mechanics-related problems. By leveraging data derived from established physical laws, machine learning models can implicitly learn the underlying physics, enabling rapid and accurate predictions.

## 1.2 Artificial neural network applications in mechanics-related domains

Over the past several decades, artificial neural network (ANN) models have seen widespread adoption across various mechanics-related domains, including solid mechanics (Kumar and Kochmann 2022), fluid mechanics (Brunton et al. 2020), and interdisciplinary domains (Guo et al. 2022). According to Janiesch et al. (Janiesch et al. 2021) and Goodfellow et al. (Goodfellow et al. 2016), an ANN is a computational model that mimics the functionalities of neurons of the human brain that can automatically learn and extract the features detection in every hidden layer. The features extracted from each layer are usually in a hierarchical way as the number of layers gets deeper. Based on the number of layers in the ANN model, the ANN algorithms could be hierarchically summarised into shallow neural networks and deep neural networks (DNN), as illustrated in **Fig. 1**.

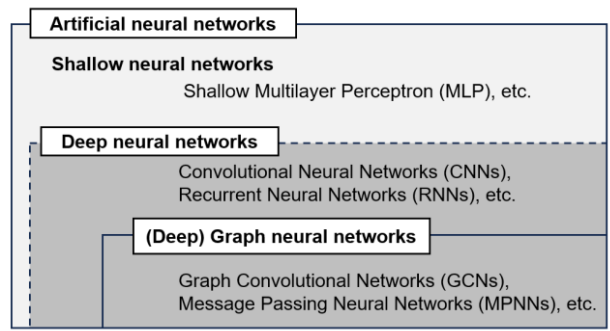

**Fig. 1** Venn diagram of ANN classifications. Typical examples for each ANN class are provided. Deep neural networks are more expressive and can model complex systems at the cost of interpretability. Therefore, the gradient of background shading (ranging from light to dark grey) symbolizes this increase in modelling power and decrease in interpretability across the ANN classes. Since there has not yet been a versatile demarcation of deep neural networks and shallow neural networks in the literature (Janiesch et al. 2021; Schmidhuber 2015), a dashed line is adopted to identify the boundary between the two classes (Adapted from Janiesch et al. (2021)).

Shallow ANNs are typically referred to as ANNs with no more than two hidden layers. Narayanasamy and Padmanabhan (2012) used an ANN model with two hidden layers to predict the spring back angle in the air-bending process. Al-Jarrah and AL-Oqla (2022) integrated a three-layer back propagation neural network and a two-layer feed-forward shallow neural network to predict the mechanical properties of green fibres. Nakamura et al. (2022) compared the abilities of linear stochastic estimation and shallow multi-layer perceptron (MLP) with a single hidden layer by considering two canonical fluid flow regression problems. They discovered that the shallow MLP is more resistant to noisy perturbations and uncertainties in data. However, due to their small number of hidden layers, the shallow ANNs are usually less capable of recognizing or modelling complex hierarchical patterns in a dataset.

DNN models overcome the abovementioned limitations by enabling more detailed feature learning through deeper hidden layers. Therefore, they are highly advantageous in handling large-scale and noisy datasets that are characteristics of many mechanics-related tasks. Deep MLP models have been adopted in domains such as damage pattern recognition (Junior et al. 2018), hydrodynamic performance predictions (Liu et al. 2022a), and milling cutting force estimation (Zuperl and Cus 2004). Zhou et al. (Zhou et al. 2019) used a deep recurrent neural network (RNN) model incorporating two long short-term memory (LSTM) layers and one bidirectional LSTM layer to identify the impact load of nonlinear structures. In recent years, convolutional neural networks (CNNs) have emerged as a valuable tool in modelling the dynamics of mechanics systems. CNNs are a type of DNN that leverages convolutional and pooling layers to exploit the spatial structure of grids or images. Consequently, they have been prevalent in building surrogate models for regressing complex mechanics-related problems. In the solid mechanics domain, CNNs have been successfully adopted to predict various physical fields, including heat transfer (Tamaddon-Jahromi et al. 2020), stress distributions (Bolandi et al. 2022b; Bolandi et al. 2022a; Kantzos et al. 2019; Khadilkar et al. 2019; Liang et al. 2018; Nie et al. 2019; Spruegel et al. 2017), material deformations (Attar et al. 2022; Attar et al. 2021a; Attar et al. 2023; Attar et al. 2021b; Xu et al. 2021; Zhou et al. 2021; Zhu and Li 2023), and topology optimisations (Attar et al. 2022; Attar et al. 2023). Meanwhile, in the fluid mechanics domain, CNNs have found vast applications in turbulence modelling (Kutz 2017), flow visualisation and analysis (Morimoto et al. 2020; Rabault et al. 2017), aerodynamics performance prediction (Afshar et al. 2019; Viquerat and HACHEM 2019; Zhang et al. 2017), flow disturbance detection and sensor optimisation (Hou et al. 2019; Le Provost et al. 2020).

However, CNNs also face limitations when addressing certain mechanics-related tasks. This is because CNNs are inherently tied to Eulerian representations with regular grids, while most solid mechanics-related tasks benefit from Lagrangian representations. For instance, modelling the deformation of cantilever beam structure under prescribed boundary conditions (Black and Najaf 2022). Moreover, CNNs are often inefficient when dealing with irregular grids, which is essential for most mechanics-related tasks. For instance, higher spatial resolution is often allocated to areas where increased precision is necessary, such as around the tip of a crack or the wing of an aerofoil. Moreover, CNNs suffer from low scalability issues. CNN models usually require fixed-sized input images because they carry out convolutional operations with fixed parameters, such as filter size and stride. Therefore, when the input images in a training dataset are of different sizes, additional image pre-processing methods are required, which can be highly time-consuming and inefficient.

Given the abovementioned limitations, there is a growing demand for alternative approaches to better handle the intricacies of mechanics-related tasks. GNNs have been introduced as a specialised type of neural network designed for processing graph-based data (Wu et al. 2020). As a brief introduction, GNNs operate by aggregating and updating node embeddings through multiple layers, with each node gathering information from its neighbours (Velivckovi'c 2023; Zhou et al. 2020; Zhou et al. 2018). This makes them highly adaptable for learning from irregular data akin to that of mechanical-related domains and providing flexibility when learning on irregular or even adaptive domain discretisation. Several GNN models have demonstrated improved generalisability to data that, while still adhering to the same underlying physical laws, may appear out-of-distribution to human observers (Black and Najaf 2022; Meyer et al. 2022; Pfaff et al. 2020; Sanchez-Gonzalez et al. 2020; Shao et al. 2023). In GNNs, graphs for mechanics-related tasks are locally connected through FE meshes and can be processed in a permutation invariant manner (Black and Najaf 2022; Pfaff et al. 2020; Sanchez-Gonzalez et al. 2020). This setup facilitates local computations, allowing the models to learn local rules that are inherently more robust to geometries outside the training regime.

### 1.3 General introduction to GNNs and their applications in various domains

#### 1.3.1 General introduction to GNNs

Sperduti and Starita (1997) pioneered the application of neural networks to graph data with an innovative approach centred on directed acyclic graphs (DAGs). Based on this motivation, Scarselli et al. (2009) proposed the first GNN model, which is capable of directly processing graphs. This allows handling a broader range of data types, such as cyclic graphs, thus representing a significant advancement over previous approaches restricted to DAGs.

Inspired by the success of CNNs, researchers have developed techniques to adapt the principle of convolution operation to graph data. Monti et al. (2017) proposed a generalised CNN architecture for non-Euclidean data, which performs convolution operations on graphs and manifolds.

The convolution operation on graph data can be performed using either spectral or spatial approaches. The spectral approaches process graph signals in the Fourier domain and performs graph convolution, whereas the spatial approaches convolve directly in the graph domain using information from the neighbourhood. The continual refinement of these methodologies has led to state-of-the-art GNNs. This paper will focus on the GNN frameworks that utilise graph convolution as a propagation module because they are the most widely used frameworks in mechanics-related domains. More details on the GNN convolution operations will be introduced in Section 0.

**1.3.2 GNN application domains**

GNNs have been used to tackle a wide range of machine-learning tasks on graph-structured data due to their unique ability to capture the intricate dependencies between nodes and edges in graphs. Typical graph learning tasks can be classified into node-level, edge-level, and graph-level tasks. Node-level tasks include node classification, node regression, and node clustering. Node prediction tasks predict the node features for unlabelled nodes; node classification involves assigning labels to individual nodes based on their attributes and the structure of their neighbourhood; node clustering aims to group nodes with analogous features or connectivity patterns together. Edge-level tasks include edge classification, determining the types of interaction between nodes, and link prediction, which predicts the likelihood of edge presenting between nodes. Graph-level tasks involve the prediction of the entire graph, including graph classification which classifies the graphs, and graph regression which assigns a numerical value to a graph. Node features predictions are the most commonly applied tasks for mechanics-related domains, whereas the other types of tasks are more frequently utilised in various other fields.

The realm of GNNs has been extensively covered in existing general review papers, offering comprehensive insights into the algorithms and architectures (Abadal et al. 2022; Bacciu et al. 2020; Georgousis et al. 2021; Veličković 2023; Zhou et al. 2022). These papers not only provided the state-of-the-art taxonomies on GNN algorithms but have also shed light on the potential future directions for GNN development. Furthermore, several in-depth review papers have delved into specific applications, showcasing representative use cases and bridging the gap between theoretical foundations and practical implementations of GNN architectures (Asif et al. 2021; Gupta et al. 2021; Liang et al. 2022; Wu et al. 2021; Zheng et al. 2022; Zhou et al. 2020). Initially, GNNs gained promise in diverse domains such as image processing, natural language processing, and speech recognition (Adamczyk 2022; Kipf and Welling 2016; Luo and Mesgarani 2019; Malekzadeh et al. 2021; Mandelli and Berretti 2022; Warey and Chakravarty 2022; Zijian Wang 2022). Then, their potential has progressively extended to engineering practices, for instance, spatial-temporal GNNs have been extensively employed in traffic forecasting applications such as traffic flow predictions (Bui et al. 2022; Jiang and Luo 2022; Ye et al. 2022). In the field of bioinformatics, the applications of GNNs in drug design and drug-target interactions have been summarised in dedicated review papers (Li et al. 2023b; Xiong et al. 2021; Yi et al. 2022; Zhang et al. 2021; Zhang et al. 2022). Moreover, GNNs have been successfully utilised in predicting the molecular properties of materials (Reiser et al. 2022; Wieder et al. 2020), leveraging the ability of graph data to represent molecules and chemical bonds as nodes and edges. The state-of-the-art approaches of GNN have also found practical implementations in internet recommender systems, with several review papers highlighting their applications and future directions in this domain (Deng 2022; Gao et al. 2023; Wang et al. 2020; Wu et al. 2023). Furthermore, review papers covering other fields of applications, such as text classification, electronic design and power systems, have also contributed to our understanding of GNNs (Liao et al. 2022; Lopera et al. 2021; Malekzadeh et al. 2021).

### 1.4 Summary and main contributions of this paper

The main objective of this review article is to provide a comprehensive survey of GNN applications in mechanics-related domains, along with identifying key challenges and potential future directions in the corresponding domains. We include a summary of recent GNN advancements in solid mechanics, fluid mechanics and interdisciplinary mechanics-related domains for readers to gain comprehensive understanding of the application of GNNs across these domains. Our main contributions are:

- Summarised and illustrated the fundamental GNN algorithms that are commonly applied in mechanics-related domains.
- Provided a review of GNN applications in different mechanics-related domains, including various graph representation methods, novel GNN models, and further discussions.
- Summarised open data and source code of GNN models from the literature, serving as a repository for researchers interested in applying GNNs in the corresponding application subdomains.

This review article is organised as follows: Section 2 reviews the preliminaries and notations of the graph theory and the fundamental algorithms of GNNs. Section 0 summarises the GNNs applications in mechanics, classified into solid mechanics, fluid mechanics and interdisciplinary mechanics-related domains. The graph representation methods, GNN models, further discussions regarding challenges and future works are discussed for each mechanics subdomain. To facilitate future research, the open data and source codes provided by the literature in section 0 are summarised in section 4. Finally, section 5 presents the conclusions and overall future directions for GNN applications in mechanics-related domains.

## 2. Fundamentals of Graph Neural Networks

This section provides an overview of the fundamental concepts related to GNNs, including conventional notations used to describe graph data and neural network building blocks. Some relevant architectures of GNN building blocks are then discussed and categorised.

## 2.1 Conventional notations for GNNs

In general, a graph $G$ can be represented by $G = (\mathbf{u}, V, E)$, where $\mathbf{u}$ is a global graph level feature, $V = \{v_1, v_2, \ldots, v_N\}$ and $E = \{e_1, e_2, \ldots, e_M\}$ denote sets of $N$ nodes and $M$ edges respectively. The adjacency matrix $\mathbf{A}_G$, which is an $n \times n$ matrix, describes the connection status of $G$. The component $A_{ij} = 1$ if nodes $v_i$ and $v_j$ are connected by an edge, otherwise $A_{ij} = 0$. For a special case when self-loops are present, i.e., an edge connecting a node to itself, the corresponding diagonal component of the adjacency matrix will be 1. In an undirected graph, $(v_i, v_j) = (v_j, v_i)$ for any edges and the adjacency matrix $\mathbf{A}_G$ is symmetric. Whereas in a directed graph, edges are orientated from one node to another and are irreversible. This adjacency matrix that is important for feature propagation in graph convolution operations which will be detailly explained in the following section. A directed edge $e_k = (v_{s_k}, v_{r_k}) \in E$ defines the edge connecting two nodes, specifically the sender node $s_k$ and the receiver node $r_k$. The set of neighbouring nodes of a node $v_i$ can be denoted by $N_G(v_i)$, and the degree of the node $d_G(v_i) = |N_G(v_i)|$ is the number of neighbouring nodes. The degree matrix $\mathbf{D}_G$ is a diagonal matrix with $D_{ii} = d_G(v_i)$, it is essential for normalisation of the feature propagation process which will be discussed in the following section. The Laplacian matrix, which is a useful measure in the spectral graph theory, can be defined as $\mathbf{L}_G = \mathbf{D}_G - \mathbf{A}_G$. This provides insightful information of the structure of a graph, for instance, the smallest eigenvalue of the Laplacian indicates how well a graph is connected. Graph data also contains node and edge features $\mathbf{x}_v \in \mathbf{R}^{p_v}$ and $\mathbf{x}_e \in \mathbf{R}^{p_e}$, where $p_v$ and $p_e$ are the dimensions of node and edge feature vectors respectively. Matrices $\mathbf{X}_V \in \mathbf{R}^{N \times p_v}$ and $\mathbf{X}_E \in \mathbf{R}^{M \times p_e}$ respectively denote the node and edge feature matrices. Within a GNN, $\mathbf{H}^{(l)}$ and $\mathbf{h}^{(l)}$ denote the hidden feature matrix and vector on the $l$th layer of the neural network, where $\mathbf{H}^{(0)} = \mathbf{X}$ is the input feature matrix.

## 2.2 Convolution operation on graph data

Although different convolution operators aggregate and propagate information in different manners, the principles of graph convolution operations are similar to conventional CNNs. Indeed, an image can be seen as a specially constrained case of graph data following Euclidean structure. Each pixel in an image can be considered a node containing node features like greyscale values or colours, and the connections between the neighbouring pixels represent the spatial relationships. During convolution operation, the features on neighbouring pixels are aggregated to the centre pixel and propagate to the next layer. Following similar principle, graph convolution operation is a more general procedure that can be applied to non-Euclidean graph data, allowing for more general applications beyond traditional image data. **Fig. 2** illustrates the difference between 2D image convolutional operation and graph convolutional operation.

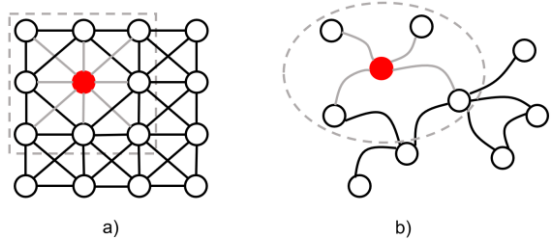

**Fig. 2** Comparison between image convolution and graph convolution a) 2D convolution operation on an image b) Graph convolution operation, for the pink node, the features of the neighbouring nodes are aggregated through the grey edges to produce the updated representation in the next layer.

The convolution operation on graph data can be divided into two categories: spectral approaches and spatial approaches. The spectral approaches involve convolving graph signals in the spectral domain based on the foundation of graph signal processing, utilising the graph Fourier transform (Shuman et al. 2013). Whereas the spatial approaches perform convolution on neighbouring nodes in the graph domain. It should be noted that the differences between the aforementioned approaches are becoming indistinct as the GNN architecture evolves. Under highly specific conditions, certain spectral filter approximations are nearly equivalent to spatial methodologies in their application and outcomes (Georgousis et al. 2021).

## 2.3 GNN building blocks

Many popular GNN frameworks adopt mechanisms similar to that of CNNs that work on images, whereby aggregation of nodal information happens in the spatial domain. Similar to CNNs, a comprehensive GNN architecture are usually composed out of multiple layers each consisting of a building block with unique weights, the following subsections describe different variations of convolution operations within a single building block.

### 2.3.1 Message passing neural network (MPNN)

One of the most widely used frameworks is the message passing neural network (MPNN) proposed by Gilmer et al. (2017). This framework contains two forward-passing phases: a message passing phase and a readout phase. The message passing phase aggregates node features along edges with a message function $M_l(\cdot)$ and the update function $U_l(\cdot)$ both with learnable parameters typically based on MLP neural networks for each edge for message passing. This spatial graph convolution operation can be denoted as:

$$\mathbf{h}_v^{(l+1)} = U_l\left(\mathbf{h}_v^{(l)}, \sum_{u \in N(v)} M_l\big(\mathbf{h}_v^{(l)}, \mathbf{h}_u^{(l)}, \mathbf{x}_{vu}^e\big)\right) \tag{1}$$

where $\mathbf{x}_{vu}^e$ is the feature vector for edge between the two nodes $v$ and $u$. The updated node hidden feature vector can either be directly passed to an output layer to perform node-level prediction or to a readout function $R(\cdot)$ based on MLPs to perform graph-level prediction:

$$\mathbf{x}_G = R\big(\mathbf{h}_v^{(L)} | v \in G\big) \tag{2}$$

where $L$ denotes the number of total layers. The message passing process is a general form of many existing graph convolution operations. The MPNN serves as a comprehensive framework incorporating a variety of GNN building blocks, with differently defined message functions, update functions and readout functions. These frameworks have been used in various applications in the field of engineering (Gao et al. 2022b; Mozaffar et al. 2021; Pfaff et al. 2020; Prachaseree and Lejeune 2022a; Sanchez-Gonzalez et al. 2020).

### 2.3.2 Graph Network (GN)

Battaglia et al. (2018) proposed the graph network (GN), one of the most general frameworks that unifies a number of existing convolution approaches. The core component of a GN is the GN block which aggregates and updates information at the node, edge and graph level. A GN block defines three update functions $\phi(\cdot)$, and three aggregation functions $\rho(\cdot)$ on the node, edge and graph levels:

$$\begin{aligned}
\mathbf{h}_{e_k}^{(l+1)} &= \phi^e\left(\mathbf{h}_{e_k}^{(l)}, \mathbf{h}_{v_{r_k}}^{(l)}, \mathbf{h}_{v_{s_k}}^{(l)}, \mathbf{u}\right), & \bar{\mathbf{h}}_{e_i}^{(l+1)} &= \rho^{e \to v}\big(\mathbf{H}_{E_i}^{(l+1)}\big) \\
\mathbf{h}_{v_i}^{(l+1)} &= \phi^v\big(\bar{\mathbf{h}}_{e_i}^{(l+1)}, \mathbf{h}_{v_i}^{(l)}, \mathbf{u}\big), & \bar{\mathbf{h}}_e^{(l+1)} &= \rho^{e \to u}\big(\mathbf{H}_E^{(l+1)}\big) \\
\mathbf{u}^{(l+1)} &= \phi^u\big(\bar{\mathbf{h}}_e^{(l+1)}, \bar{\mathbf{h}}_v^{(l+1)}, \mathbf{u}\big), & \bar{\mathbf{h}}_v^{(l+1)} &= \rho^{v \to u}\big(\mathbf{H}_V^{(l+1)}\big)
\end{aligned} \tag{3}$$

The update functions $\phi^e$, $\phi^v$ and $\phi^u$ update the node, edge and graph features from one layer ($l$) to the next ($l + 1$); a local aggregation function $\rho^{e \to v}$ aggregates the updated edge features to each node; two global aggregation functions $\rho^{e \to u}$ and $\rho^{v \to u}$ aggregate the updated edge and node features at the graph level. In general, the aggregation functions should be invariant to permutations of nodes and edges.

Diverse combinations of the activation and aggregation functions define many variations of the GNN building blocks. The MPNN can be seen as a variant of the GN without global processing functions.

In practice, most applications of GN are in the form of Encoder-Processer-Decoder architecture consisting of multiple GN blocks. The encoder part transforms the features into a latent space, which are then subject to a sequence of processor GN blocks. The decoder GN blocks are similar to the encoder part but operate inversely to convert the latent vectors back into the ultimate form of output data.

### 2.3.3 Graph convolutional network (GCN)

As mentioned before, graph convolution can be operated in both the spatial and spectral domains. The spectral approaches process graph signals in the spectral domain by making use of the graph Fourier transform, specifically, by computing the eigen-decomposition of the graph Laplacian. A typical example of the spectral approaches is the Chebyshev spectral CNN (ChebNet) (Defferrard et al. 2016). ChebNet approximated the spectral graph kernel with the K-localised Chebyshev polynomial of the eigenvalue matrix of the Laplacian matrix. Spectral graph convolution computation can be expensive due to the eigen-decomposition of the Laplacian matrix, especially for large graphs.

Motivated by the ChebNet, Kipf and Welling (2016) proposed the graph convolutional network (GCN) by employing the first-order approximation of the Chebyshev polynomial. The propagation module can be described in matrix form as:

$$\mathbf{H}_V^{(l+1)} = f\left(\widetilde{\mathbf{D}}_G^{-\left(\frac{1}{2}\right)} \widetilde{\mathbf{A}}_G \widetilde{\mathbf{D}}_G^{-\left(\frac{1}{2}\right)} \mathbf{H}_V^{(l)} \mathbf{\Theta}\right) \tag{4}$$

where $l$ is the layer number, $f(\cdot)$ is the activation function, $\widetilde{\mathbf{A}}_G = \mathbf{A}_G + \mathbf{I}_N$, and $\widetilde{\mathbf{D}}_G = \text{diag}\left(\sum_{k=1}^{N} \widetilde{\mathbf{A}}[1,k], \ldots, \widetilde{\mathbf{A}}[N,k]\right)$, $\mathbf{\Theta}$ is a matrix filled with learnable parameters. Although this localised feature aggregation approach is derived from the graph Laplacian's spectral decomposition, the simplification avoids the need for full eigen-decomposition. The GCN framework applies convolutions by aggregating features from the immediate neighbours of each node. This aggregation is a localised operation, which is a characteristic of spatial methods. Therefore, in practice, GCNs can be seen as a hybrid leveraging the theoretical foundations of spectral approaches while implementing convolution in a spatial-like approach. Due to this characteristic, GCNs can be seen as specific cases of the MPNN and GN which operate without considering the edge features.

### 2.3.4 Graph attention network (GAT)

While the aforementioned frameworks assume identical contributions of neighbouring nodes, several approaches have been proposed to perform spatial graph convolution based on different weighting strategies. For example, the graph attention network (GAT) Veličković et al. (2017) adopts an attention mechanism. The GAT utilises a self-attention mechanism which computes the weights of each node spatially beside the target node, and the attentional convolution operation can be denoted as:

$$\mathbf{h}_v^{(l+1)} = \sigma\left(\sum_{u \in N(v)} a_{vu} \mathbf{W}^{(l+1)} \mathbf{h}_u^{(l)}\right) \quad (5)$$

where $\sigma$ is the sigmoid activation function, the attention weight $a_{vu}$ is implicitly computed by neural networks to determine the importance of each neighbour to the node, $\mathbf{W}^{(l+1)}$ is the weight matrix of the $l+1$ layer.

### 2.3.5 GraphSAGE

A number of GNN frameworks encounter a challenge in the form of heterogeneous degrees of node connectivity, leading to a non-uniform distribution of neighbourhood sizes. This gives rise to computational inefficiencies when attempting to incorporate the entire node's neighbourhood during graph convolution operations on large graphs. This is because storing and processing the connectivity information of nodes with a high degree of connections requires substantial memory resources. In addition, the complexity of aggregating features from differently sized neighbourhoods slows down processing and influences computational efficiency, especially when normalising contributions from neighbours. Specially tailored aggregation functions need to be defined for different types of graph data in terms of graph sizes and connection types.

Hamilton et al. (2017) introduced an inductive representation learning technique on large graphs, namely the GraphSAGE (Sample and Aggregate). GraphSAGE implements a sampling method to obtain a uniform number of neighbouring nodes for every node in the graph. The graph convolution operation can be defined by:

$$\mathbf{h}_v^{(l+1)} = \sigma\left(\mathbf{W}^{(l+1)} g_{l+1}\left(\mathbf{h}_v^{(l)}, \left\{\mathbf{h}_u^{(l)} \forall u \in S_{\mathcal{N}(v)}\right\}\right)\right) \quad (6)$$

Where $S_{\mathcal{N}(v)}$ is a fixed sized sample of the neighbourhood of node $v$,. The aggregation function $g$ can be either mean, sum or max aggregator.

All the aforementioned GNN frameworks have been wide adopted in various mechanics-related applications. Based on the characteristics of each scenario, appropriate frameworks are selected to optimally address the specific challenges.

## 3. GNNs applied to various mechanics-related domains

This section summarises the GNN applications in mechanics-related domains based on existing papers from peer-reviewed journals, conferences, and preprints. According to section 1.1, the mechanics-related applications of interest in this review paper are categorised into solid mechanics, fluid mechanics, and interdisciplinary mechanics-related domains. Such a categorisation scheme is based on unique challenges and commonalities of each domain, necessitating unique GNN models to address these challenges, thereby ensuring clarity in analysis and ease of reference for domain-focused readers. We particularly emphasise the utilisation of GNNs in solid mechanics applications, as most of their applications are reliant on Lagrangian representations which GNNs excel in.

### 3.1 Application of GNNs in the solid mechanics domain

This section presents a comprehensive review of GNN applications in solid mechanics domain with a detailed summary illustrated in **Table 1**. These applications were summarised into three main subdomains, including continuum and frame-based structures, mechanical metamaterials, and fracture mechanics and tool wear predictions. For each subdomain, we reviewed the graph representation methods, GNN models and discuss the challenges and future works of the GNN applications.

**Table 1** GNN application summary in the solid mechanics domain.

| Applications | Descriptions | GNN frameworks | References |
|---|---|---|---|
| Continuum and frame-based structures | Displacement field prediction; 2D elastostatic problem of cantilever beam | Multi-fidelity MPNN with subgraph analysis and physics-informed training | Black and Najaf (2022) |
| | Displacement field prediction; 2D or 3D linear elastic hollow cylinder | GCN with physics-informed loss functions | Gao et al. (2022a) |
| | Displacement field prediction; 3D Clothes deformation in contact with a rigid ball | Multigraph MPNN with learned adaptive remeshing | Pfaff et al. (2020) |
| | Displacement field prediction; 3D Hyper-elastic plate | Spatial-temporal multigraph MPNN | Pfaff et al. (2020) |
| | Displacement field prediction; 3D Elastic plate subject to an actuator | Multiscale MPNN | Cao et al. (2022) |
| | Displacement field prediction; 2D/3D beam | GCN model with input-independent learnable weighted multi-channel aggregation | Deshpande et al. (2022) |
| | Displacement field prediction; 3D linear elastic and hyper-elastic beam | GCN with physics-informed training | He et al. (2023) |
| | Stress field prediction; 2D cantilever beam | GCN with attention modules | Fu et al. (2023) |
| | Structural dominant failure modes classification; 2D truss structure | MPNN with attention modules | Tian et al. (2024) |
| | Dynamic responses prediction; 3D elastic beam and 2D elastic frame | GN combined with RNN with physics-informed training | Chen et al. (2024) |
| | Stiffness prediction; 3D post milled aircraft structural parts | GCN with graph level features | Chen et al. (2022) |
| | Buckling direction prediction; 2D asymmetric buckling columns | GCN | Prachaseree and Lejeune (2022b) |
| | Structural design; 2D beam layout design | MPNN with GraphSAGE | Zhao et al. (2023) |
| | Structural design; 3D geometry feature designs | MPNN | Wong et al. (2022) |
| Mechanical metamaterials | Wave propagation and displacement prediction; 2D shell metamaterial | MPNN with physics-based edge update function | Xue et al. (2023) |
| | Dominant deformation mechanism classification; 3D truss metamaterial | MPNN classification model | Indurkar et al. (2022) |
| | Deflection prediction; 3D truss metamaterial | GCN | Ross and Hambleton (2021) |
| | Stiffness prediction; 3D shell metamaterial | MPNN | Meyer et al. (2022) |
| | Metamaterial design; 2D architected materials | Semi-supervised GCN | Guo and Buehler (2020) |
| | Generative modeling and inverse design; 3D truss metamaterial | Graph based VAE combined with a property predictor | Zheng et al. (2023) |
| | Metamaterial inverse design; 2D truss metamaterial | GCN (edgeconv) and MPNN | Dold and van Egmond (2023) |
| Fracture mechanics and tool wear predictions | Fracture mechanics crack propagation prediction; 2D brittle materials | Spatial-temporal MPNN | Perera et al. (2022) |
| | Crack field and displacement field prediction | Multi-scale MPNN | Perera and Agrawal (2024) |
| | Tool wear field prediction; 3D York metal forging die | GCN | Shivaditya et al. (2022) |

### 3.1.1 Continuum and frame-based structures

The continuum and frame-based structures subdomain emphasises the study of macroscopic structures such as beams, columns, plates, and frames, and how these structures respond to external forces and moments. Typical GNN applications in this subdomain include predicting the deformation fields (Black and Najaf 2022; Deshpande et al. 2022; Gao et al. 2022a; He et al. 2023; Pfaff et al. 2020), stress fields (Fu et al. 2023), and the structural dominant failure mode of beam or frame structures under various loading conditions such as point loads or contact between objects (Tian et al. 2024).

**Graph representation methods**

The graph representation methods can be categorised into continuum and frame-based structures, as illustrated in **Fig. 3**. The graph construction methods for continuum structures are based on FE meshes. The FE mesh nodes represent graph nodes, and the FE mesh edges represent graph edges. Input node features include external nodal forces (Deshpande et al. 2022; Fu et al. 2023; Tian et al. 2024), nodal coordinates (Fu et al. 2023; He et al. 2023; Pfaff et al. 2020; Tian et al. 2024), material properties (Fu et al. 2023), and node types for heterogeneous graphs (Pfaff et al. 2020). Input edge features include the distance between each node (Black and Najaf 2022; Pfaff et al. 2020; Tian et al. 2024), and edge types represent different boundary conditions (Fu et al. 2023). Additionally, it is worth mentioning that node positional coordinates can differ under different global coordinate systems. Therefore, the GNN model trained with data from one global coordinate system might not be applicable to other coordinate systems. To apply a coordinate-free GNN model, a recent study from Chen et al. (2024) used the relative nodal displacements in node attributes and the change of edge length as edge features. Additionally, they also included gravitational acceleration and timestep interval as graph level features (Chen et al. 2024). While for the frame-based structure, graph nodes could represent joints or connection points in the frame, graph edges could represent the discrete elements connecting these graph nodes as shown in **Fig. 3(b).** Geometric positions and constraints could be modelled as graph node features while material properties, length of the edge could be embedded in graph edge features.

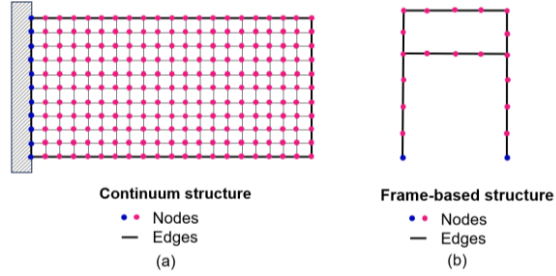

**Fig. 3** Common graph representation methods for (a) Continuum and (b) frame-based structures.

Zhao et al. (2023) extensively studied the graph representation of frame-based structures based on different node and edge types. Three different graph representation methods named Case-Normal, Case-Inverse, and Case-AllNodes were adopted, as shown in **Fig. 4**. The Case-Normal method represented columns as graph nodes and represented both beam and non-beam links as graph edges. The Case-Inverse method represented columns as graph edges and represents beam and non-beam links as graph nodes. The Case-AllNodes method described the columns, beam links, and non-beam links all as graph nodes. They discovered that the Case-Inverse method achieved the best overall accuracy compared to other graph representation methods.

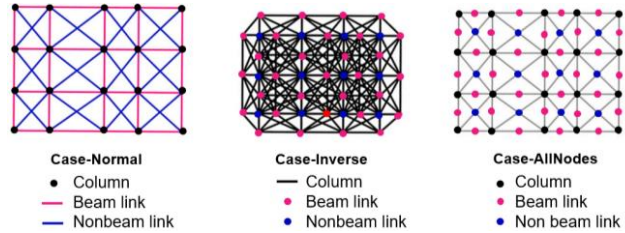

**Fig. 4** Various graph representation methods explored for frame structure. (Adapted from Zhao et al. (2023))

However, in Finite Element (FE) simulations, stress and strain states are computed on mesh elements rather than mesh nodes. Some case studies with the primary aim of predicting the stress and strain fields in the mesh elements adopted a novel graph representation method by encoding information of the abovementioned node and edge features in a condensed element level node. Fu et al. (2023) proposed a novel graph embedding approach named Boundary-oriented Graph Embedding (BOGE). In BOGE, the information of mesh nodes, edges, and elements was combined to form a condensed graph representation, as illustrated in **Fig. 5**. In the condensed graph representation, the graph nodes were the central nodes of mesh elements, and graph edges represented the connectivity between neighbouring mesh elements. Notably, instead of one-hop neighbouring edges, shortcuts were established between the target mesh element with both the global boundary mesh elements and local multi-hop mesh elements, as illustrated in **Fig. 5**. Therefore, fewer message-passing steps were required to accurately predict boundary value problems with long-range graph-vertex interactions.

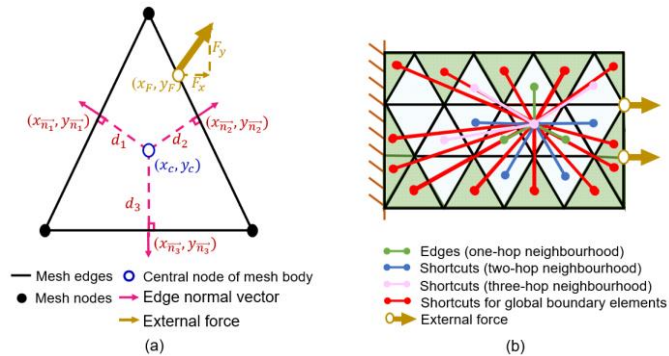

**Fig. 5** Graph representation methods proposed by Fu et al. (2023) (a) Condensed graph representation features for one triangular mesh element (b) An illustration for shortcut between the target mesh element with multi-hop neighbours and boundary elements. (Adapted from Fu et al. (2023))

Additionally, external loadings may not always be explicitly defined in FE simulations for direct inclusion in graph inputs. Instead, they can be implicitly represented based on the contact interactions between multiple objects. Pfaff et al. (2020)

proposed a multigraph MPNN architecture named MeshGraphNet (MGN) that can implicitly learn mesh-based self-collisions or contact between two objects. The graph structure in MGN had the addition of world graph edges that were generated based on a predefined hyperparameter of connectivity radius, as displayed in **Fig. 6**. Therefore, the graph structure in MGN was referred to as a multigraph. The multigraph represented node features as nodal coordinates in the material mesh and global world spaces. It also represented edge features as nodal coordinate difference and Euclidean distance in both the material mesh space and 3D cartesian world space (Pfaff et al. 2020). The mesh space focused more on message passing in the material space. In contrast, the world space analysis focused more on message passing in the world space involving contact between objects and self-collisions. Additionally, they trained a dynamic and a sizing field model to perform learned dynamic re-meshing during rollout (Pfaff et al. 2020). All learned re-meshing variants could adeptly adjust the mesh resolution during each test time step and were, therefore, able to provide improved performance to the GNN model without the need to include a domain-specific re-mesher in the GNN model.

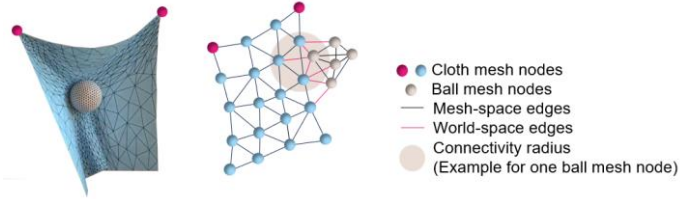

**Fig. 6** Graph representation method for modelling the contact between objects. (Adapted from Pfaff et al. (2020))

Moreover, the abovementioned graph construction methods often concern high-fidelity analysis with high-resolution mesh structures. A high-fidelity mesh may contain tens of thousands of mesh edges and mesh nodes with multiple channels of edge or node features. Thus, storing the nodes and edges features and training the GNN models can be highly time-consuming. Therefore, an analysis using a multi-fidelity GNN (MFGNN) is illustrated in **Fig. 7** (Black and Najaf 2022). The MFGNN model took the inputs of FE simulation results of a low-fidelity mesh and outputs high-fidelity predictions. The input of the MFGNN model included a low-fidelity graph with simulation results assigned as node and edge features. A high-fidelity graph with unassigned node and edge features is also included in the graph input. The node and edge features in the low-fidelity graph were projected to the high-fidelity node and edge features. The high-fidelity graph was then subdivided into several subgraphs to update nodes and edge features in these subgraphs using message passing functions. Finally, the output of the high-fidelity graph was retained from these subgraphs via a feature recovery process. After training, Black and Najaf (2022) discovered that the MFGNN with subgraph analysis could accurately predict the displacement field near boundary conditions. This may be because the subgraph analysis function in the MFGNN model enabled more localised analysis within each subgraph to capture more specific physical relations near the boundary conditions (Black and Najaf 2022).

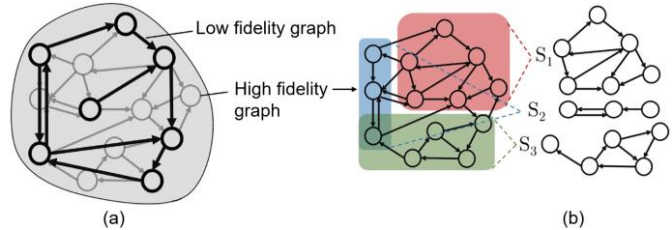

**Fig. 7** The multi-fidelity graph representation method and subgraph analysis proposed by Black and Najaf (2022) (a) Illustration of low and high-fidelity graph (b) The high-fidelity graph is separated into three subgraphs (S1, S2, S3) and analysed separately. (Adapted from Black and Najaf (2022))

**GNN algorithms**

According to **Table 1**, MPNN with edge-based feature processing is popular (Black and Najaf 2022; Pfaff et al. 2020; Tian et al. 2024). GCN models are also widely adopted (Deshpande et al. 2022; Gao et al. 2022a; He et al. 2023; Zhao et al. 2023). Some continuum and frame-based structures related GNN applications have extended from these fundamental architectures of MPNN and GCN, as described in Section 2.3, based on specific problems they aim to address.

To further enhance the GNN message passing capabilities in learning long-range dependencies of geometric features and boundary conditions, a multi-scale graph learning approach was developed by Deshpande et al. (2022) named Multi-channel Aggregation Network (MAgNET). The MAgNET framework is an encoder-decoder architecture inspired by the CNN U-Net. During encoding, one or more Multichannel Aggregation layers (MAg) were applied to the mesh-based dataset, followed by a pooling layer to downsample the graph. The aggregation and pooling processes were repeated until a desired coarse mesh level was obtained. In the decoding stage, unpooling operations on the graph data were performed and followed by one or more MAg layers. Compared to CNN U-Net benchmark results, the graph-based method was claimed to have improved scalability by operating on unstructured data (Deshpande et al. 2022).

Moreover, physics-informed training is widely adopted to improve prediction accuracy. For instance, an internal potential energy-based loss function was adopted in the training process to model the elastostatic problem of a cantilever beam (Black and Najaf 2022), and a deep energy method was adopted to model the linear elastic and hyperelastic problems of a cantilever beam (He et al. 2023). The deep energy method is a method to solve physical problems by minimising the total energy instead of directly dealing with the PDEs, such as elasticity equations for linear elastic cantilever beam problems (He et al. 2023). Additionally, Chen et al. (2024) incorporated the rest states such as the internal forces of elements in a continuum deformable body as the initial hidden states of the gated recurrent unit (GRU) when developing a combined RNN and GNN surrogate model. They composed the loss function based on the predicted nodal acceleration and internal forces as these two criteria were essential to compute in the second-order differential equations governed by Newton's second law. It was discovered that the RNN architecture helped to reduce the accumulated long-term rollout error since it was trained to reduce the loss of multi-step sequence prediction during model training. Furthermore, it is worth noting that Gao et al. (2022a) utilised GCN model with Galerkin variational formulation of physics-informed loss functions. They achieved more accurate prediction results compared with CNN baselines and have faster convergence compared to finite volume-based iterative solvers.

Attention-based GNN methods are also incorporated to facilitate a more nuanced understanding of graph structure and message passing processes. Tian et al. (2024) developed a hierarchical GAT with node and edge attention modules built inside the MPNN fundamental architecture. Compared to the genetic algorithm (GA) and the β-unzipping method, the proposed attention-based GNN algorithm exhibited higher computational efficiency.

**Further discussions**

Most GNN applications in this domain often exclude material properties from their input tensors and have minimal material variations in their training dataset. The absence of explicit material featurisation in the input tensor means that even though the model might implicitly learn the properties of specific materials used in the training set, its generalisation performance is limited when encountering new materials, thus constraining the capability of GNNs to learn and generalise across diverse material characteristics. This limitation likely stems from the challenges encountered in dataset preparation, particularly the difficulty of assembling datasets that represent a variety of material properties alongside complex and high-dimensional geometric parameterisation schemes and requires future investigations. Moreover, most GNN models focus on learning linear material behaviours such as linear elastic materials. Furthermore, the majority of GNN models are primarily designed to capture nonlinear dynamics under the assumption of linear material behaviours, such as that exhibited by elastic materials. However, their effectiveness in accurately modelling nonlinear material behaviours, such as those involving elastoplastic, remains limited and requires further investigation.

GNN performance is sensitive to the size of input variables and their physical meanings (Fu et al. 2023). If very few input variables are embedded, the GNN model may tend to underfit due to a lack of effective physical embeddings. However, if the input variables size is too large, the GNN model may become highly computationally intensive and prone to overfitting. Therefore, future work is expected to investigate various input combinations of the GNN model to explore the potential of optimal input combinations.

Furthermore, the current models mainly focus on relatively small graphs with several thousands of nodes and edges involved. When modelling the dynamics of complex continuum structures, finer meshes are required to observe the localised features such as displacement or stress concentrations. This would result in significant increase of nodes and edges numbers in the graph embeddings. The training time may be considerable due to additional edges and nodes required to represent the system accurately. Although previous methods such as MFGNN is proposed to address this issue, the MFGNN is not an independent solver as it requires FE simulations to obtain low fidelity graph embeddings.

Moreover, when operating on large graphs, the message-passing layers may get deeper, resulting in gradient vanishing or gradient explosion (Lukovnikov et al. 2020). Therefore, novel GNN architectures addressing this issue could be adopted. For instance, Addanki et al. (2021) explored a GNN model for deep transudative node classification that leverages bootstrapping, and a very deep inductive graph regressor capable of handling up to 50 layers regularised by denoising techniques. Conversely, if message-passing layers are relatively shallow, the GNN model may lack the capacity to capture the long-range dependency in the system. For instance, the deformation of the cantilever beam next to the boundary wall may be influenced by the force applied to the cantilever beam at the other end. With shallow layers, the message propagation may not be able to capture this type of long-range dependencies. To mitigate this problem, multi-scale GNNs with down-sampling techniques can be explored in a hierarchical order by enhancing the message passing over graphs (Cao et al. 2022).

### 3.1.2 Mechanical metamaterials

Mechanical Metamaterials (MMs) are micro or nano-scale materials designed with a hierarchical architecture (Surjadi et al. 2019). MMs exhibit unique mechanical properties such as high strength-to-density ratio and high resilience. Studies have extensively focused on predicting the deformation mechanisms (Indurkar et al. 2022; Xue et al. 2023) or other structure-property relations, such as elastic moduli and heat conductivity of MMs (Meyer et al. 2022). Additionally, it is worth noting that some inverse designs of MMs with targeted desired properties have emerged in recent years (Dold and van Egmond 2023; Zheng et al. 2023).

**Graph representation methods**

Developing effective graph construction methods for MMs is essential to enhance input featurisation, which is crucial for the learning and performance of GNN models. For truss-based metamaterials, common graph construction methods include representing the nodes in the 3D metamaterial structure as graph nodes and the struts as graph edges, as shown in **Fig. 8(a)**. Node features often include nodal positions and load levels, while edge information is commonly edge length or strut orientations (Indurkar et al. 2022). Moreover, Ross and Hambleton (2021) encoded node types into the graph system to learn the boundary representation of the periodic unit cell structure.

Besides truss-based metamaterials, a shell-based metamaterial can be represented by a cross-spring system that contains rigid crosses connected by nonlinear springs, as illustrated by **Fig. 8(b)**. In the cross-spring system, the rigid crosses are represented as graph nodes, and the nonlinear springs are encoded as graph edges. The nodal features contain the position and orientation of the rigid elements, and the edge features contain elastic energy stored in each spring (Xue et al. 2023).

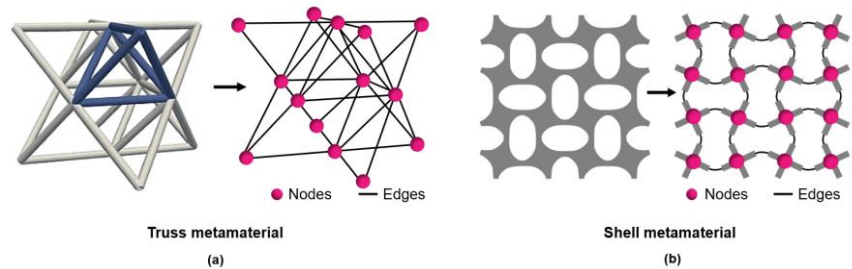

**Fig. 8** Examples of graph representation methods for (a) truss metamaterial and (b) shell metamaterial. (Adapted from Zheng et al. (2023) and Xue et al. (2023))

**GNN algorithms**

From **Table 1**, commonly applied GNN algorithms include MPNN and GCNs. In addition, several unique GNN architectures with tailored functions have been proposed (Guo and Buehler 2020; Xue et al. 2023). To improve prediction accuracy, Xue et al. (2023) incorporated a physics-based edge update function based on the functional form of elastic energy of the cross-spring system. This approach provided a quantitatively accurate solution to predict the dynamics of large-scale structures comprising over $200 \times 200$ metamaterial unit cells with hyper-elastic properties. Guo and Buehler (2020) adopted a semi-supervised learning approach to classify test node load levels to obtain the load levels of architecture metamaterials. The input graph contained load level data for only 1% of graph nodes, and the trained model was able to accurately predict the load levels for the remaining 99% of graph nodes. Notably, this model incorporated a predefined design approach that allowed for the creation of voids in low-load areas while maintaining the overall volume of the material. This approach ensured that in predicted high-load areas, the edge stiffness would increase, approaching a maximum stiffness value. Conversely, in predicted low-load areas, the edge stiffness would decrease, potentially reaching zero.

Most GNN applications in MMs mainly focus on forward prediction tasks such as predicting the structural-property relations. Regarding inverse design, the GNN architecture is often coupled with design algorithms using gradient descent. Moreover, some generative models were also proposed to generate various MMs with different geometric features. Zheng et al. (2023) developed a graph variational autoencoder-based generative model that enables continuous latent space exploration encompassing a vast array of truss metamaterials. The generative model also enabled the convenient generation of new designs via latent space traversal or structure interpolation. Based on the generative model, they further developed an optimisation framework for inverse truss design, targeting custom mechanical properties, including exceptionally stiff, auxetic, pentamode-like, and tailored nonlinear behaviours (Zheng et al. 2023). Moreover, Dold and van Egmond (2023) developed an approach that utilises differentiable message-passing algorithms with EdgeConv layers in GCNs. EdgeConv updated node features by considering the differences between neighbouring nodes and applying linear transformations followed by max-pooling. Additionally, they integrated a GRU to iteratively update node features. They compared the GNN model to an MPNN baseline and gradient descent optimisation method to permit modifications to the geometric configurations and local attributes of individual lattice elements. They discovered that the EdgeConv surrogate model had enhanced material property prediction accuracy compared to the MPNN model.

**Further discussions**

In practical applications in mechanical metamaterials, visco-hyperelastic materials are often used (Xue et al. 2023). While Xue et al. (2023) investigated the dynamic response of a beam structure composed of shell-based metamaterials with a simplified hyperplastic behaviour, they did not account for viscosity and energy dissipation. To capture the rate-dependant behaviour due to viscosity, RNNs can be utilised to capture temporal variations or MGN networks can be employed to learn material or dynamic constitutive equations, facilitating accurate predictions of system dynamics (Xue et al. 2023). Additionally, when investigating the dynamic response of complex structures consisting of a large number of unit cell metamaterials, issues may arise due to the increasing number of unit cells and their long-range dependencies. Long-range dependency in metamaterials refers to the influence that the behaviour or state of one unit cell can have on distant unit cells, beyond the immediate vicinity. To mitigate this, advanced neural network architectures, such as hierarchical GNN models or networks with established shortcuts between boundary conditions and local unit cells, could be explored to enhance the model capability to learn dynamic responses on a global scale rather than local scale. However, how to effectively down-sample the model without losing information on the unit cell representation of the metamaterial is also a challenging topic to discover.

Finally, most applications focus on the forward prediction of metamaterial properties given a predefined structure, while very few explore the inverse design of metamaterials. The metamaterial structure is determined by various design parameters that can vary independently. Consequently, the solution to an optimal design is not unique, often due to the presence of local optima in gradient-based optimisation methods. Addressing this non-uniqueness is critical for the success of the inverse design problem, requiring the development of robust optimisation strategies to effectively navigate the complex design space.

### 3.1.3 Fracture mechanics and tool wear predictions

Fracture mechanics and tool wear predictions involve studying the propagation of defects in materials, such as predicting tool wear conditions of manufacturing tooling, which is essential in solid mechanics applications to predict material or structural failures. Although there has yet to be much research going on in this domain, the current GNN applications in this subdomain include modelling the crack propagation in brittle materials (Perera and Agrawal 2024; Perera et al. 2022), and tool wear field prediction of manufacturing tools (Shivaditya et al. 2022).

**Graph representation methods**

Regarding GNN applications in crack propagation prediction, crack tips could be represented by graph nodes, and graph edges could represent the connectivity between multiple crack tips within a zone of influence. Node features may include cartesian coordinate positions and orientations of the crack-tip (Perera et al. 2022). Meanwhile, for tool wear field prediction, the FE mesh of the manufacturing tooling could be constructed into graphs with mesh nodes and mesh edges representing graph nodes and graph edges, respectively. The node features may include temperature and friction coefficient, while the edge information may include the connectivity between node pairs (Shivaditya et al. 2022).

**GNN algorithms**

When modelling crack propagation, spatial-temporal MPNN could be adopted to predict crack propagation of multiple time steps. Due to the intricacies of microcrack mechanics, Perera et al. (2022) proposed that the MicrocrackGNN framework consisted of four sub-GNN surrogate models, each predicting an underlying physical component of microcrack mechanics. The first two GNN sub-models predicted Mode-I and Mode-II stress intensity factors, the third GNN sub-model predicted crack propagation type, and the final GNN sub-model predicted the position of microcrack tips. The trained MicrocrackGNN demonstrated its capability to simulate crack propagation across various initial microcrack configurations, ranging from 5 to 19 microcracks. Notably, the prediction speed of the MicrocrackGNN was 25 times faster than traditional finite element simulations (Perera et al. 2022). Moreover, compared to CNN-RNN baseline models Recurrent Convolutional Neural Network (RCNN) and Recurrent Encoder–Decoder Neural Network (REDNN), the sub-GNN surrogate models exhibited an order of magnitude smaller error relative to the baseline. Specifically, the RCNN model, which includes convolutional layers, batch normalisation layers, ReLU activation functions, and a linear output layer, is designed to handle sequential data. In contrast, the REDNN model employed an encoder-decoder architecture with convolutional and transpose convolutional layers, allowing it to reconstruct input data for enhanced feature learning. The results showed that the MicrocrackGNN outperformed both RCNN and REDNN models, highlighting its superior accuracy and efficiency in predicting crack propagation. However, they discovered that conventional mesh-based GNNs often face oversmoothing issues, especially when dealing with large graphs with high-fidelity meshes. The oversmoothing issue was due to insufficient message passing capabilities. To address this, Perera and Agrawal (2024) further introduced a multiscale GNN framework with adaptive mesh refinement method. This framework reduced the number of required message-passing steps by sequentially coarsening and upscaling the mesh. The trained framework demonstrated faster simulation while maintaining high accuracy across all cases compared to traditional physics-based phase-field fracture models (Perera and Agrawal 2024).

Regarding tool wear predictions, Shivaditya et al. (2022) introduced a GCN architecture to predict the wear field of metal forging tools. In the GCN surrogate model, the first GCN layer takes a graph with 5 node features and outputs a graph with 50 hidden node features. The second GCN layer processed the graph into 100 hidden node features, followed by a ReLU activation layer. The subsequent three GCN layers reduce the hidden node features from 100 to 50 and then to 1, targeting the wear feature. Dropout layers were applied for regularisation after each activation function. The final ReLU layer ensures that all model outputs are positive. They discovered that the GCN surrogate model was at least four times more accurate compared to other benchmark models, such as Point-Net (Qi et al. 2017) and the Dynamic Graph Convolutional Neural Network (DGCNN) model (Wang et al. 2019).

**Further discussions**

The development of GNN applications in fracture mechanics is still in its nascent stage. The applications primarily focused on relatively fewer number of cracks; it is suggested that future work could extend from the MicrocrackGNN architecture mentioned in (Perera et al. 2022) and study the prediction performance on larger numbers of initial microcracks. Additionally, when modelling spatial-temporal MPNN architectures, the error accumulation in long-term rollout may influence model prediction accuracy. Integrating RNN with MPNN may help reduce the error accumulation since RNNs are trained to reduce the loss of multi-step sequence prediction. Regarding tool wear field prediction, Shivaditya et al. (2022) mentioned that additional features, such as normal stress and flow stress, have yet to be included in the GNN model to reduce computational effort. More case studies to analyse different input features and their relations with wear field prediction results may be beneficial. Finally, it is advised that the generalisability performance of GNN models in the fracture mechanics and tool wear prediction subdomain could be evaluated, which is essential for industrial practices. Moreover, future work would be beneficial in expanding these GNN applications to damage mechanics such as predicting the dislocation in the crystal structure of a material.

### 3.2 Application of GNNs in the fluid mechanics domain

This section summarises various GNN applications within the fluid mechanics domain. In this section, the applications were categorised into hydrodynamics, aerodynamics, and complex fluid rheology subdomains. In the hydrodynamics subdomain, the fluid of focus is usually liquids such as water or oil. Therefore, it is widely applied to tasks such as hydraulic machinery, submarine design, and ocean simulations. Aerodynamics is often concerned with air or other types of gaseous fluids. Studying the principles of aerodynamics is crucial for the design of vehicles and for optimising sporting facilities. Complexity fluid rheology often involves complex fluids that do not follow Newton's law of viscosity. This fluid type is known as non-Newtonian fluid, including polymers, suspensions, and biological fluids like blood. The complexity of fluid rheology is often seen in its applications across various industries, including cosmetics, food, and biomedicine. In line with the previously outlined categorisation methods, the graph representations, GNN algorithms, and challenges and future works in each fluid mechanics subdomain are briefly discussed in this section. A comprehensive summary of GNN applications in the fluid mechanics domain is given in **Table 2**.

**Table 2** GNN application summary in fluid mechanics domain.

| Applications | Descriptions | GNN frameworks | References |
|---|---|---|---|
| Hydrodynamics | Eulerian method; Ocean, river hydrodynamics | GCN with multiscale graph representation | Shi et al. (2022) |
| | | Multiscale spatial-temporal MPNN with adaptive remeshing | Lino et al. (2022a) |
| | Lagrangian method; Fluid fall and box bath | Spatial-temporal MPNN | Sanchez-Gonzalez et al. (2020) |
| | | GCN | Li and Farimani (2022) |
| | Lagrangian method; 3D watercourse-inverse design | Spatial-temporal MPNN | Allen et al. (2022b) |
| Aerodynamics | Eulerian method; Incompressible laminar flow over cylinder | GCN with GAT | Liu et al. (2022b) |
| | | GCN with ResBlock module for node aggregation, physics-informed training, and attention mechanism | He et al. (2022) |
| | | MPNN with multiscale and graph level representation | Yang et al. (2022) |
| | | Spatial-temporal MPNN | Li et al. (2023c) |
| | Eulerian method; Incompressible laminar flow over aerofoil | GCN with GraphSAGE | Ogoke et al. (2021) |
| | | Spatial-temporal MPNN | Gao et al. (2022b) |
| | | Multiscale spatial-temporal MPNN | Fortunato et al. (2022) |
| | | GCN combined with CFD | Belbute-Peres et al. (2020) |
| | | GCN | Jessica et al. (2023) |
| | Eulerian method; Incompressible laminar flow over arbitrary shapes | GCN | Chen et al. (2021) |
| | | Rotational equivariant multiscale MPNN | Lino et al. (2022b) |
| | Eulerian method; Fluid field prediction and optimisation of turbomachinery | GCN | Li et al. (2023a) |
| | Eulerian method; Turbulence modelling of the wake in flow over a cylinder | MPNN with physics informed training | Quattromini et al. (2023) |

| | Eulerian method; 3D turbulent flow predictions of urban wind fields | Spatial-temporal MPNN with physics-informed training | Shao et al. (2023) |
| | Eulerian method; 2D turbulent channel flow | MPNN | Dupuy et al. (2023) |
| | Lagrangian method; 3D laminar Taylor-Green Vortex, and 3D reverse Poiseuille Flow | Rotation equivariant spatial-temporal MPNN | Toshev et al. (2023) |
| Rheology of complex fluid | Eulerian method; Blood, yogurt, or polymer solutions | Rheology-informed multi-fidelity GNN | Mahmoudabadbozchelou et al. (2022) |

### 3.2.1 Hydrodynamics

Applications of GNNs in the hydrodynamics subdomain have been concerned with large graph predictions of temperature fields and fluid flow of oceans or rivers (Lino et al. 2022a; Shi et al. 2022). Additionally, various rollout tests of fluid particles interacting with each other inside a constrained space, such as the dynamics of fluid particles in water cubes and bathtubs (Allen et al. 2022b; Li and Farimani 2022; Sanchez-Gonzalez et al. 2020), are popular in this subdomain.

**Graph representation methods**

We have further divided the GNN applications in fluid mechanics subdomains based on the Eulerian and Lagrangian methods to provide a more precise distinction and navigation to readers interested in either method. The Eulerian method uses a fixed grid system to analyse fluid motion (Harlow and Hirt 1972). Through finite-difference methods, the mass, momentum, and energy conservation laws are applied to the fixed grids to calculate fluid behaviours. For instance, the 2D cylinder flow problem uses the Eulerian method because the mesh grid for the control domain is fixed. In contrast, the Lagrangian method studies fluid dynamics by tracking the movement of individual fluid particles over time (Wang 2023). It involves applying the mass, momentum, and energy conservation laws to a mesh of particles that move along with the fluid.

Because of the abovementioned distinct methods in analysing fluid behaviours, there are significant differences in graph construction techniques when applying GNN with Eulerian and Lagrangian methods. In the Eulerian method, people are commonly interested in the changes in physical properties over time at fixed spatial points. Therefore, the graph is usually static. In the static graph, each node is fixed in spatial location, and the interaction with its neighbourhood is constant (Shi et al. 2022). In the Lagrangian method, fluid particles were encoded as graph nodes and graph edges were added between particles within a connectivity radius (Lino et al. 2022a). The graph edges representing the current interactions between particles are also dynamic to accommodate changes in graph node interactions over time (Lino et al. 2022a; Sanchez-Gonzalez et al. 2020) (Li and Farimani 2022). For instance, two fluid particles previously close to each other and interacting strongly may move relatively far away in the next time step, causing interactions between them to weaken or disappear.

Additionally, it is noticed that storing intermediate feature maps at full resolution can be challenging because of limited GPU storage space. To solve this problem, Shi et al. (2022) used graph hierarchical tree cutting to build a graph hierarchy consisting of graphs at different resolution levels. Nevertheless, modelling hydrodynamics of oceans and rivers often requires effective message passing on large graphs with an excessive number of message-passing layers. Propagating the node and edge attributes between nodes separated by hundreds of hops would require an extensive number of message passing layers, which could be inefficient and unscalable (Lino et al. 2022a). Therefore, Lino et al. (2022a) proposed multiscale MPNN that created graphs with different levels. To illustrate, lower-resolution graph levels possess fewer nodes and edges and, therefore, require a smaller number of messages passing layers to propagate node and edge features over longer distances effectively.

**GNN algorithms**

Commonly applied GNN algorithms in the hydrodynamics domain are spatial-temporal MPNNs to model the evolution of fluid properties such as velocity and pressure field at incremental timesteps, mimicking the physics simulator (Sanchez-Gonzalez et al. 2020). A few GCN models also emerged (Li and Farimani 2022; Shi et al. 2022). To accurately predict the fluid motion with a far less parameterised model, Li and Farimani (2022) decomposed a single forward step in the fluid dynamics into multiple simpler sub-systems, namely node-focused network and edge-focused network. The node-focused network is a GCN with distance-weighted aggregation methods, and the edge-focused network could be regarded as a GCN with only edge information.

**Further discussions**

Overall, the GNN applications in hydrodynamics have showcased their improved training efficiency without compromising the accuracy of prediction results. Additionally, improved generalisability performance was discovered in the MPNN network (Sanchez-Gonzalez et al. 2020). However, the current spatial-temporal MPNN model is trained on next-step prediction and unrolled with a fixed time step (Sanchez-Gonzalez et al. 2020). Future work could potentially analyse adaptive time stepping, investigate sequence models and the role of training noise in error accumulations, or incorporate an RNN-based model for reducing error accumulations (Lino et al. 2022a).

### 3.2.2 Aerodynamics

From **Table 2**, GNN applications in the aerodynamics subdomain mainly involve predicting 2D aerofoil flow, cylinder flow, or flow over arbitrary shapes. The prediction outputs are primarily fluid field predictions such as velocity field, pressure field, and wall shear stress. Moreover, there are several GNN applications in the 3D domain, such as the 3D flow field in turbomachinery (Li et al. 2023a), turbulence modelling of wake region over a cylinder (Quattromini et al. 2023), turbulent flow predictions of urban wind fields (Shao et al. 2023), turbulent channel flow (Dupuy et al. 2023), as well as 3D laminar Taylor-Green Vortex and reverse Poiseuille Flow (Toshev et al. 2023).

**Graph representation methods**

Most GNN applications of aerodynamics problems adopted the Eulerian method, especially when studying laminar fluid dynamics. Meshes in CFD can be naturally interpreted as graphs, where mesh nodes are represented as graph nodes and mesh edges are represented as graph edges. Node input features often include nodal coordinates, while edge features include node connectivity information and the relative distance between connecting node pairs (Belbute-Peres et al. 2020; Ogoke et al. 2021; Pfaff et al. 2020; Shao et al. 2023). Moreover, Toshev et al. (2023) included the external force field of the reverse Poiseuille flow as a node feature when studying 3D turbulent flow.

In addition to the abovementioned graph representation methods, a graph augmented representation method based on node-element hypergraph is proposed by Gao et al. (2022b), as illustrated in **Fig. 9**. In the data representation stage, the node-element hypergraph encoded mesh nodes as graph nodes. The graph edges were constructed based on element-node edges, relative locations between element centres and nodes were encoded in the edge feature. The proposed graph representation method produced more stable and accurate results compared to the conventional graph construction method for a longer rollout time step within the training range (Gao et al. 2022b).

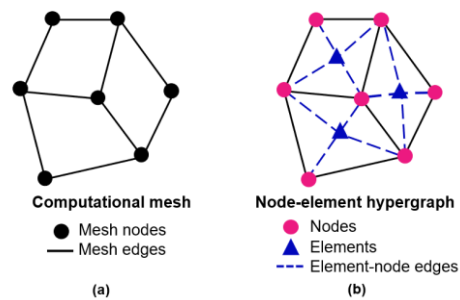

**Fig. 9** Comparison of (a) computational mesh and (b) node-element hypergraph. (Adapted from Gao et al. (2022b))

Moreover, Lino et al. (2022b) applied a rotation-invariant representation of the input and output vector fields with directed angles between adjacent edges. By detailing the angles between edges and incorporating the lengths of these edges, this approach offers a comprehensive representation of the relative positions of nodes without explicitly requiring a coordinate system. This rotational equivariant data enabled higher prediction accuracy than the explicit representation method based on coordinate systems, which directly uses the coordinates of the nodes to define their positions.

**GNN algorithms**

MPNN and GCN have been applied to fluid dynamics problems, as in **Table 2**. Spatial-temporal MPNN models are commonly adopted (Gao et al. 2022b; Shao et al. 2023; Toshev et al. 2023) to perform rollout tests that predict intermediate time steps rather than end-to-end predictions. Spatial-temporal MPNN models allow the GNN model to capture the complexity of flow evolutions such as turbulence and vortex formation. Additionally, several examples have studied multiscale MPNNs to improve message-passing capabilities (Fortunato et al. 2022; Lino et al. 2022b; Yang et al. 2022). Fortunato et al. (2022) proposed the Multiscale MGN as a hierarchical update of the MGN. The multiscale MGN model transferred 'high accuracy' labels from a high resolution simulation onto a coarser, lower resolution mesh, achieving more accurate and less time-consuming prediction results than the MGN using the same number of message passing layers (Fortunato et al. 2022). While this model contained only two levels of graph hierarchies, Shao et al. (2023) and Toshev et al. (2023) explored more levels in the graph hierarchies. Shao et al. (2023) designed a Multiscale GNN to deal with multiscale graph data following a CNN U-Net-inspired encoder-decoder architecture. Toshev et al. (2023) developed an algebraic multigrid-based coarsening layer that coarsened the graph to smaller scales. They then summarised and extracted features through message passing at smaller graph scales. Finally, a graph recovery layer was applied to recover the coarse graph to the high-fidelity graph of the initial mesh scale.

Regarding GCN applications, attention-based GCN models were commonly adopted (He et al. 2022; Liu et al. 2022b). He et al. (2022) developed a GCN model consisting three graph convolution layers and two spatial gradient attention layers with a ResBlock module and a ResSE module. Additionally, Liu et al. (2022b) used a multi-head attention mechanism to learn node relations in different representation subspaces. Experimental results showed that the ability of the attention mechanism to aggregate physical information of adjacent nodes was much better than that of average aggregation (Liu et al. 2022b). Belbute-Peres et al. (2020) developed a CFD-GCN model that integrated a CFD solver with the GCN architecture. The CFD-GCN model operates on two types of graphs: a fine graph and a coarsened version of the fine graph. The coarsened graph was processed with a CDF simulator, while the fine mesh was processed simultaneously by GCNs. The results from the coarsed mesh simulation results were then upsampled and combined with the intermediate outputted from the fine graph. Additional layers of GCNs were then applied to the combined features, to output the final desired output values such as velocity and pressure fields at each node in the fine graph. In the CFD-GCN framework, the CFD simulator named SU2 can be considered an additional 'layer' attached to the GCN architecture. An interface was built to enable the CFD solver to be integrated seamlessly as any other layer within the PyTorch module.(Belbute-Peres et al. 2020). The results demonstrated that the CFD-GCN framework can improve the accuracy of CFD simulation results on the coarsened mesh (Belbute-Peres et al. 2020). Moreover, they discovered enhanced generalisability when compared to the conventional GCN baseline without the upsampling technique. CFD-GCN was able to approximate the characteristics of the unseen shockwave behaviour.

Moreover, some of these existing GNN-based models applied the laws of physics to the flow field prediction to further improve prediction accuracy. He et al. (2022) incorporated physics-based training utilising Navier–Stokes loss functions, and Quattromini et al. (2023) ensured that the predictions result satisfied the Reynolds-Averaged Navier-Stokes (RANS) equations, which were fundamental in fluid dynamics simulations.

**Further discussions**

It is noticed that current studies are mainly based on 2D laminar flow. Future work is expected to study large 3D simulations using GNNs as they often provide more accurate and realistic fluid dynamics results that satisfy industrial applications. For instance, 3D models can simulate wingtip vortices and induced drag, providing a more accurate and realistic analysis. Also, experimental fluid simulation results utilising particle image velocimetry (PIV) or Digital image correlation (DIC) methods to track the movement of fluid particles could be adopted as a training dataset for GNNs to learn from experimental data.

Moreover, many GNN applications in the aerodynamics domain are predicting spatial-temporal properties of fluid fields. Although these incremental GNN models can capture the dynamics of mechanical systems, the inherent challenge arises regarding accumulated time-step errors, especially in long-term rollouts. At training time, the spatial-temporal MPNN network never sees unphysical states, such as the non-zero divergence in incompressible Navier-Stokes equations. However, those unphysical states can appear through error accumulation. One method to mitigate error accumulation includes introducing training noise, such as Gaussian noise, to perturb the input data. However, this approach often has to compensate for relatively lower prediction accuracy and needs to be more principled. Therefore, a better understanding of rollout error accumulations could be studied, and the role of training noise in error accumulations could be further explored. Combining RNNs with GNN models may be an alternative approach that reduces the accumulated error.

### 3.2.3 Rheology of complex fluid

Another fluid mechanics problem was the rheological behaviour of complex fluids. Mahmoudabadbozchelou et al. (2022) proposed the rheology-informed graph neural networks (RhiGNets) which is capable of learning the stress response of a complex fluid with applied deformation through a limited number of experiments. In the RhiGNets model, the graph nodes represent particles, and the graph edges represent the interactions or forces between particles. The architecture of the RhiGNets model comprises three sections, encoder, decoder, and graph predictor. It can efficiently achieve accurate results for predicting the shear, elongation, or oscillation of complex fluids such as blood, yoghurt, or polymer solutions.

### 3.3 Application of GNNs in interdisciplinary mechanics-related domain

The interdisciplinary mechanics-related GNN applications are thoroughly reviewed in this section. We have categorised the literature into interdisciplinary mechanics-related domains by their commonalities into thermo-mechanics, biomechanics, fluid-structural interaction, and complex system dynamics subdomains. The thermo-mechanics subdomain involves applications that study the interaction between mechanical and thermal effects in mechanical systems. The biomechanics subdomain integrates mechanics and biology principles to effectively understand the mechanical behaviours of biological systems. The fluid structural interactions subdomain studies the coupling effects of deformable structures and fluids. Lastly, the complex system subdomain is categorised by multiple interacting systems with nonlinear dynamics. These systems frequently display chaotic dynamics, making predictions particularly challenging.

GNNs are exceptionally well-suited for modelling the intricate, interconnected, and nonlinear nature of the high-dimensional systems. There is a growing interest in adopting GNN applications to understand, predict, and control mechanical systems within these subdomains. A summary of the GNN applications in these subdomains can be found in **Table 3**. This section concisely discusses the data representations, GNN architecture, and model performance of each proposed GNN framework in these subdomains.

Table 3 GNN application summary in interdisciplinary mechanics-related domain.

| Applications | Descriptions | GNN frameworks | References |
|---|---|---|---|
| Thermo-mechanics | Thermal profile prediction; Additive manufactured objects | MPNN combined with RNN | Mozaffar et al. (2021) |
| | Thermal behaviour prediction; 3D electronics objects | Spatial-temporal MPNN | Sanchis-Alepuz and Stipsitz (2022) |
| Biomechanics | Tissue deformation prediction; Liver and brain | MPNN with physics-informed training (liver) | Dalton et al. (2023) |
| | | GCN GraphSAGE and GraphConv (brain) | Salehi and Giannacopoulos (2022) |
| | Cardiac mechanics prediction; Left ventricle of the heart | MPNN (MGN and DeepEmulator) | Dalton et al. (2021) |
| | | Multiscale MPNN | Dalton et al. (2022) |
| | | MPNN with physics-informed training | Dalton et al. (2023) |
| Fluid-structure interaction | Dynamics prediction; Flag blown by the wind | Spatial-temporal multigraph MPNN | Pfaff et al. (2020) |
| Complex system dynamics | Dynamics Inference for complex systems such as ropes, bouncing balls, and splashing fluids | Spatial-temporal MPNN | Sanchez-Gonzalez et al. (2018) |
| | | Constraint-based spatial-temporal MPNN | Rubanova et al. (2021) |
| | Motion Inference of multiple rigid bodies colliding with one another | Spatial-temporal MPNN | Allen et al. (2022a) |
| | Complex 3D granular flow | Spatial-temporal MPNN | Mayr et al. (2021) |

### 3.3.1 Thermo-mechanics

The existing studies in the thermos-mechanics subdomain are mainly based on predicting spatial-temporal thermal responses for additive manufacturing processes (Mozaffar et al. 2021; Xue et al. 2022).

**Graph representation methods**

For additive manufacturing applications, graph structures commonly represent FE meshes, with graph nodes representing mesh nodes and graph edges representing mesh edges. Node features included node types that indicated whether a node is an active node, layer height that represents the distance from the bottom of the substrate, and the inverse of the distance between nodes and the laser beam. Edge information was stored in the adjacency matrix, with distances between nodes encoded as edge features.

**GNN algorithms**

Sanchis-Alepuz and Stipsitz (2022) studied the spatial-temporal thermal behaviour of 3D electronics designs such as heat sinks and printed circuit boards using a spatial-temporal MPNN architecture. They discovered that the MPNN model was more computationally efficient than traditional FE simulations. However, they also discovered that although the one-step prediction error was very low, the accumulated error of the MPNN surrogate model increased significantly (Sanchis-Alepuz and Stipsitz 2022). To mitigate these accumulated errors in rollout tests, Mozaffar et al. (2021) proposed a recurrent graph neural network (RGNN) to model the spatial-temporal thermal profile of the laser powder bed fusion (LPBF) additive manufacturing process. The RGNN surrogate model comprised DeeperGCN layers, GRU modules that took the update of the DeeperGCN layer and a hidden stage as input to predict the temperature field at incremental test time. DeeperGCN proposed by Li et al. (2020) employed a residual-based design with skip connections, which empowered deeper and more efficient training of the GCN model compared to conventional GCN architectures. The performance of RGNN was compared to the conventional GNN framework. Results revealed that the RGNN model excelled in longer-term predictions of over 1,000 time steps compared to the baseline GNN architecture.

**Further discussions**

The abovementioned GNN applications in the thermo-mechanics domain are all based on simulation results. With the continuous development of high-quality shared repositories of additive manufacturing experimental data, it is promising that more GNN applications could be developed by incorporating the experimental data to enhance the prediction accuracy and become more practical for real-world usage.

### 3.3.2 Biomechanics

Current applications in the biomechanics domain are focused on high-fidelity soft tissue deformations such as the brain (Salehi and Giannacopoulos 2022), liver (Dalton et al. 2023) and emulating the cardiac mechanics of the left ventricle of the heart (Dalton et al. 2022; Dalton et al. 2021).

## Graph representation methods

The graph representation method for the biomechanics domain is mainly based on high-fidelity meshes. The graph nodes represent mesh nodes, graph edges represent mesh edges. Node features commonly include node types indicating boundary nodes and kinematic nodes subject to deformations, nodal coordinates, and physical properties. Edge features often include distance between two nodes (Dalton et al. 2021; Salehi and Giannacopoulos 2022). Moreover, to ensure the model input is independent of different coordinate systems, Dalton et al. (2023) used the unit vector in the fibre direction instead of nodal coordinates as node features. The geometric information was mainly stored in the edge features by incorporating the relative positions (Dalton et al. 2023). However, many message-passing steps may be required to propagate information within a dense set of FE nodes. To allow for more efficient message passing, Dalton et al. (2022) introduced a multi-scale data augmentation method to generate additional layers of coarse, virtual nodes that allow rapid message passing using a 'shortcut' through the augmented multi-scale spaces as illustrated in **Fig. 10**. Additional virtual nodes were generated based on the k-means algorithm. Corresponding edges were added between the real and virtual nodes. As shown in **Fig. 10** below, the black squared nodes are real nodes from the original FE meshes, and the red circular nodes are virtual nodes created based on the positions of the real nodes. The green triangular nodes are the next graph hierarchy level of virtual nodes created based on the red circular virtual nodes (Dalton et al. 2022). **Fig. 10(b-e)** shows the additional edges generated based on the virtual nodes and original nodes.

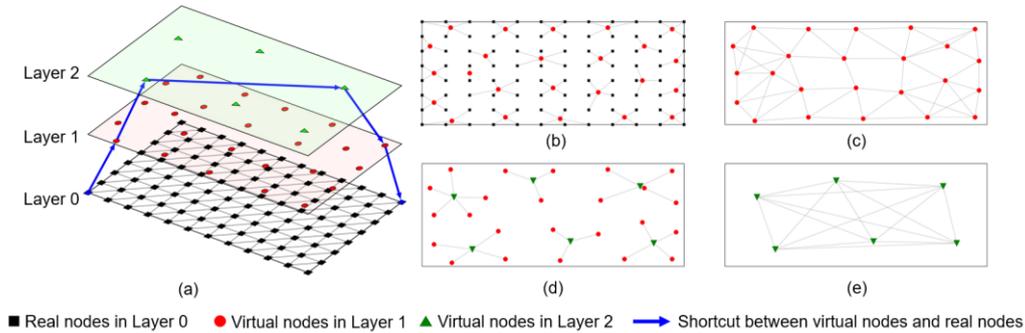

**Fig. 10** A graph representation method proposed by Dalton et al. (2022) (a) Illustration of the hierarchical graph levels generated by adding virtual graph nodes and additional node elements. Blue arrows show how message passing is achieved based on the hierarchical graphs (b-e) Illustration of the addition of virtual nodes and the generation of different graph hierarchies (layers). (Adapted from Dalton et al. (2022))

## GNN algorithms

According to **Table 3**, MPNN and GCN models are the common GNN algorithms adopted by the biomechanics subdomain applications. Regarding MPNN-based architectures, Dalton et al. (2021) conducted a comparison study between the DeepEmulator and the MGN to investigate their performances in emulating the cardiac mechanics of the left ventricle (LV) of the heart. The DeepEmulator framework is a GNN framework proposed by Zheng et al. (2021), initially developed for the application of skinning-based animations of 3D characters. They discovered that the MGN model was two times more accurate compared to the DeepEmulator model. Driven by this discovery, Dalton et al. (2022) proposed a GNN framework named DeepGraphEmulator. The DeepGraphEmulator model adopted a similar encoder-processor-decoder architecture to MGN (Pfaff et al. 2020), but with an enhanced multi-scale data augmentation method (as previously introduced in **Fig. 10**). They discovered that the proposed DeepGraphEmulator could generalise to different systems of various LV geometries and material constitutive properties, making it promising for real-world practices. To further enhance the prediction accuracy, they combined physics-informed training with the proposed DeepGraphEmulator architecture by incorporating a potential energy function in the loss function (Dalton et al. 2023).

Regarding GCN models, Salehi and Giannacopoulos (2022) proposed the PhysGNN surrogate model based on combining two GCN modules, including graphCONV and graphSAGE, with Jumping Knowledge connections. The results showed that the proposed PhysGNN surrogate model could achieve an absolute displacement error of under 1mm for 97% of the graph nodes, which satisfies the precision requirement in neurosurgery (Salehi and Giannacopoulos 2022).

## Further discussions

Firstly, one significant challenge lies in accurately representing the unique muscle fibre arrangements in the LV of different patients. Each patient's LV geometry will have a unique myofibre field, which significantly impacts the mechanical response of the LV (Dalton et al. 2022). The current GNN models for cardiac mechanics often assume uniform material properties across all nodes. However, this assumption does not accurately reflect the variations in myofibre fields among different patients. It also does not account for regional differences of LV within a single patient, especially in cases of myocardial infarction (MI), where the myocardium exhibits higher stiffness levels in the affected regions (Dalton et al. 2022). To address this, future work should allow stiffness values to vary across different regions of the LV by inputting local values in the decoder stage rather than using a global value. Additionally, existing GNN models primarily focus on the passive phase of the cardiac cycle and neglect the active phase involving myocardial contraction. To provide a more comprehensive simulation of cardiac mechanics, future research should extend the modelling to include the active phase (Dalton et al. 2022).

Incorporating physics-based training in biomechanics is a relatively new topic requiring further exploration. The commonly applied strain energy density functions of the myocardium are considered slightly compressive (Dalton et al. 2023). To comprehensively address incompressibility, future work should incorporate multi-field variational principles, introducing additional variables to enforce the incompressibility constraint (Dalton et al. 2023). Additionally, preliminary studies have shown that physics-informed GNNs can match FE results across different mesh densities, suggesting that increasing mesh density can lead to more accurate and reliable results (Dalton et al. 2023). However, constructing and managing large graphs with increased mesh density can be resource-intensive, and this requires future research on optimising the GNN models and graph representation methods to achieve better computational efficiency.

### 3.3.3 Fluid-structure interaction

One spatial-temporal MPNN architecture named the MGN, mentioned in section 3.1.1, was also adopted in various interdisciplinary mechanics-related domains, such as fluid-structure interactions (Pfaff et al. 2020).

**Graph representation methods**

The fundamental graph representation method is similar to those explained in section 3.1.1. Additionally, Pfaff et al. (2020) also developed a learned adaptive re-meshing method across different time steps. This method involves learning a sizing field from the FE simulations that specifies the desired local resolution. The GNN model dynamically adjusts the mesh resolution during iterative rollouts by splitting, collapsing, and flipping edges based on the predicted sizing field at each timestep. This ensures a finer mesh resolution in areas with rapid changes or high gradients, such as regions with high stress in the deformed material, and a coarser mesh resolution in areas with low stress.

**GNN algorithms**

A comprehensive explanation of the MGN framework can be found in section 3.1.1. The proposed model was applied to predict the evolution of fluid-structure interactions with the learned adaptive re-meshing method to improve prediction accuracy in small-scale features. It is noted that the MGN model had a robust capability to accurately predict out-of-distribution testing datasets. For example, when trained on flat flag geometries containing 2,000 nodes, the model demonstrated commendable adaptability by successfully extending their understanding to unseen cylindrical flag geometries comprising 20,000 nodes under identical air excitation conditions.

**Further discussions**

Regarding future work directions, the learned adaptive re-meshing method could be further investigated. Currently, the GNN surrogate model learns the sizing field based on training data generated by FE adaptive re-meshing algorithms. These methods are designed to improve numerical approximations of physical processes by increasing grid density in areas requiring higher resolution. However, GNNs excel in identifying and learning from structural and relational patterns within graph data, which is a fundamentally different approach from FE analysis. The FE adaptive mesh refinement method is tailored for numerical precision in physical simulations but may not optimise the graph structure for GNN processing and analysis. Therefore, a potential future direction is to explore optimal adaptive re-meshing methodologies specifically tailored for GNNs. A GNN-specific adaptive mesh might focus on optimising the graph topology to improve the efficiency of message passing and the prediction accuracy of the GNN model, rather than mimicking the mesh refinement methodologies used in numerical simulations. This could involve dynamic adjustments that prioritise connectivity and information flow within the graph, which are critical for GNN performance.

### 3.3.4 Complex systems dynamics

According to **Table 3**, GNN applications in complex systems dynamics primarily involve the dynamics inference for rigid body systems, including the movement of robot arm and pendulum (Sanchez-Gonzalez et al. 2018), contact between multiple rigid bodies (Allen et al. 2022a), movement of bouncing balls (Rubanova et al. 2021), and evolution of complex 3D granular flow (Mayr et al. 2021).

**Graph representation methods**

In these examples, the graph representation varied based on different shapes of geometries. For instance, the bodies and joints were depicted as graph nodes and edges when modelling the pendulum movement (Sanchez-Gonzalez et al. 2018). Additionally, accurately representing the boundary conditions when modelling the contact between multi-objects is crucial to ensure accurate predictions. Common approaches to model contacts between multi-objects are mainly based on node to node interactions; however, this may not always be effective, especially when the contact point occurs on mesh faces far from nodes. To compensate for this, Mayr et al. (2021) proposed Boundary Graph Neural Networks (BGNNs) that dynamically alter the graph structure by adding virtual graph nodes or graph edges at the boundary surface to minimise the distance between the virtual graph nodes and the real fluid particle, to accurately depict the contact between real fluid particles and boundary surfaces. It was demonstrated that the BGNN model could accurately predict the motion of material particles over hundreds of time steps without applying physical constraints (Mayr et al. 2021).

Additionally, although incorporating additional nodes and edges in the augmented graph can improve accuracy, it also comes with high computational costs. To accurately depict the boundary conditions during collision without adding more nodes and edges, Allen et al. (2022a) introduced the Face Interaction Graph Network (FIGNet) that processed inputs on mesh faces rather than mesh nodes. The graph construction method was similar to the MGN (Pfaff et al. 2020), but a major modification was applied in the encoding stage. During the encoding stage, Allen et al. (2022a) replaced node-to-node 'world edges' with face-to-face edges connecting the sender to the receiver face. By doing so, the FIGNet model outperformed the MGN regarding accuracy and computational efficiency.

**GNN algorithm**

Spatial-temporal MPNNs were commonly adopted for complex system dynamics to perform rollout tests (Allen et al. 2022a; Mayr et al. 2021; Rubanova et al. 2021; Sanchez-Gonzalez et al. 2018). In these GNN models, the encoder-processor-decoder architecture was employed by combining the node features from both the source and target nodes with the original edge features. The updated edge features are aggregated and combined with the target node features to output the updated node features. In the decoder layer, the node or graph-level features are concatenated through a fully connected neural network model to output final node prediction values.

Rubanova et al. (2021) introduced an MPNN-based physics simulator utilising an implicit constraint-based approach named C-GNS. This approach ensured that physical constraints were rigorously satisfied by integrating a scalar constraint function into an iterative optimisation process. The architecture employed an encoder-processor-decoder architecture based on the Graph Network-based Simulators (GNS) models proposed by Sanchez-Gonzalez et al. (2020). The GNS can be considered a predecessor of the MGN but was primarily designed for particle-based simulations, such as simulating the movement of fluid particles in a container. Compared to the GNS model, the C-GNS model had an additional scalar constraint function which quantified the degree to which a proposed state update violated physical constraints. The constraint function was employed within an iterative optimisation loop. The process began with an initial guess for the node positions. At each iteration, the model adjusted the node positions using the gradient of the constraint function with respect to the positions. The C-GNS model outperformed the GNS baseline models in terms of prediction accuracy and capability to generalise to novel and hand-designed constraints in various simulation scenarios.

**Further discussions**

It is noticed that the GNN applications in complex system dynamics were mainly based on examples with a relatively small number of nodes and edges. Future work could expand the dataset to incorporate diverse and larger graphs of more complex interactions.

## 4. Open data and source code

The open data and source codes for the applications of GNN models in the mechanics-related domains are organised in **Table 4** to assist readers with similar research interests in reproducing previous studies and promote communication and future research on potential fields. These open data and source codes were originally provided by the authors of the papers discussed in Section 0.

**Table 4** Summary table of open data and/or source codes[a].

| Descriptions | GNN frameworks | Open data and/or source codes | References |
|---|---|---|---|
| **Solid mechanics applications** | | | |
| Displacement field prediction: 2D elastostatic problem of cantilever beam | Multi-fidelity MPNN with subgraph analysis and physics-informed training | github.com/MCMB-Lab/MFGNNs | Black and Najaf (2022) |
| Displacement field prediction: 3D linear elastic and hyper-elastic beam | GCN with physics-informed training | github.com/Jasiuk-Research-Group | He et al. (2023) |
| Displacement field prediction: 3D Elastic plate subject to an actuator | Multiscale MPNN | github.com/Eydcao/BSMS-GNN | Cao et al. (2022) |
| Displacement field prediction: 2D/3D beam | GCN model with input-independent learnable weighted multi-channel aggregation | github.com/saurabhdeshpande93/convolution-aggregation-attention | Deshpande et al. (2022) |
| Displacement field prediction: 3D Hyper-elastic plate | Spatial-temporal multigraph MPNN | github.com/google-deepmind/deepmind-research/tree/master/meshgraphnets | Pfaff et al. (2020) |
| Wave propagation prediction: 2D shell metamaterial | MPNN with physics-based edge update function | github.com/tianjuxue/gnmm | Xue et al. (2023) |

[a] Accessed 20 June 2024

| Generative modeling and inverse design: 3D truss metamaterial | Graph based VAE combined with a property predictor | zenodo.org/records/8255658 | Zheng et al. (2023) |
| --- | --- | --- | --- |
| Crack field and displacement field prediction | Multi-scale MPNN | github.com/rperera12/Adaptive-mesh-based-Multiscale-Graph-Neural-Network | Perera and Agrawal (2024) |
| **Fluid mechanics applications** | | | |
| Eulerian method—Ocean, river simulations | Multiscale spatial-temporal MPNN with adaptive remeshing | github.com/afansi/multiscalegnn | Lino et al. (2022a) |
| | GCN with multiscale graph representation | github.com/trainsn/GNN-Surrogate | Shi et al. (2022) |
| Eulerian method: Incompressible laminar flow over aerofoil | Spatial-temporal MPNN | github.com/garrygale/NodeElementMessagePassing | Gao et al. (2022b) |
| | GCN with GraphSAGE | github.com/BaratiLab/Airfoil-GCNN | Ogoke et al. (2021) |
| | GCN combined with CFD | github.com/locuslab/cfd-gcn | Belbute-Peres et al. (2020) |
| Eulerian method: Incompressible laminar flow over cylinder | MPNN with multiscale and graph level representation | github.com/baoshiaijhin/amgnet | Yang et al. (2022) |
| | Spatial-temporal MPNN | github.com/Litianyu141/My-CODE | Li et al. (2023c) |
| Eulerian method: Incompressible laminar flow over arbitrary shapes | GCN | github.com/jviquerat/gnn_laminar_flow | Chen et al. (2021) |
| Lagrangian method: Fluid fall and box bath | GCN | github.com/BaratiLab/FGN | Li and Farimani (2022) |
| | Spatial-temporal MPNN | github.com/deepmind/deepmind-research/tree/master/learning_to_simulate | Sanchez-Gonzalez et al. (2020) |
| **Interdisciplinary mechanics-related applications** | | | |
| Soft tissue deformation prediction | MPNN with physics-informed training | github.com/YasminSalehi/PhysGNN | Salehi and Giannacopoulos (2022) |
| Thermal profile prediction: Additive manufactured objects | MPNN combined with RNN | github.com/AMPL-NU/Graph_AM_Modeling | Mozaffar et al. (2021) |
| Complex 3D granular flow prediction | Spatial-temporal MPNN | github.com/ml-jku/bgnn | Mayr et al. (2021) |
| Dynamics prediction: flag blown by the wind | Spatial-temporal multigraph MPNN | github.com/google-deepmind/deepmind-research/tree/master/meshgraphnets | Pfaff et al. (2020) |

## 5. Conclusions

To conclude, GNNs have gained significant traction in mechanics-related domains due to their unique ability to learn from graph-structured data. This review paper summarised various fundamental GNN algorithms, such as MPNN, GN, GCN, GAT, and GraphSAGE, and their applications in various domains. We systematically reviewed graph representation methods and GNN architectures across different mechanics-related subfield, specifically solid mechanics, fluid mechanics, and interdisciplinary mechanics-related subfields. In the solid mechanics subfield, we categorised GNN applications into continuum and frame structure, mechanical metamaterials, and fracture mechanics and tool wear predictions subdomains. In the fluid mechanics subfield, we covered hydrodynamics, aerodynamics, and rheology subdomains. For the interdisciplinary mechanics-related subfield, we explored thermo-mechanics, biomechanics, fluid-structural interaction, and complex system dynamics.

Our review highlights that in solid mechanics, most GNN applications focus on predicting displacement or stress fields of continuum structures, and structural response of mechanical metamaterials. Few studies have explored fracture mechanics, which represents a potential area for further research. In fluid mechanics, GNN applications are predominantly centred on 2D laminar flow, with fewer studies addressing turbulent modelling and rheology, both of which could benefit from more research. Compared to the solid mechanics and fluid dynamics domains, interdisciplinary mechanics-related fields that include multi-physics coupling are relatively underexplored. This is likely due to the complexity of problem formulation and the increased non-linearity that challenges GNN performance in these domains. However, with the continuous development of GNN architectures and novel graph construction methods, there is significant potential for applying GNNs in these areas, making them worth exploring in future research.

Most of the reviewed GNN models outperformed traditional DNN baselines in terms of prediction accuracy and scalability. Additionally, some case studies demonstrated the generalisability potential of GNNs, showing their ability to handle previously unseen testing data that exhibit similar physical behaviours to the training dataset. However, challenges that hinder the practical deployment of GNN models in the mechanics-related domains still need to be addressed. Therefore, challenges and future works tailored to specific types of mechanics-related problems have been discussed in each subdomain. Common challenges in these subdomains include defining graph structures and mapping features more effectively, improving message-passing efficiencies in high-fidelity graphs, embedding material non-linearity, rollout for different size of timesteps. Additionally, addressing time series error accumulations, incorporating experimental data into the training loop to enhance prediction accuracy, embedding physics-based constraints without oversimplification, and investigating generalisability for more complex, high-dimensional, and large-scale real-world applications are essential areas for future research.

Overall, this review paper compiles recent advancements at the intersection of graph theory and fundamental mechanics principles. It offers insights and a comprehensive overview for researchers looking to explore the capabilities of GNNs in addressing complex physics and mechanics-related problems.

**Declarations**

**Author contributions** Y.Z. and H.L. wrote the main manuscript, prepared figures, reviewed and revised the manuscript. H.Z., H.R.A., and T.P. reviewed the manuscript and provided valuable insights for editing. N.L. supervised and organised the work, reviewed and provided critical insights for editing the drafts.

**Funding** Partial financial support was received from EPSRC for the CASE Conversion DTP training grant (EP/R513052/1), UKRI Impact Acceleration Accounts (EP/X52556X/1), and Innovate UK smart grant (10083425).

**Conflct of interest** The authors have no competing interests to declare that are relevant to the content of this article.